# Multi-View Constrained Clustering with an Incomplete Mapping Between Views

Eric Eaton[1], Marie desJardins[2], Sara Jacob[3]

[1] Bryn Mawr College, Computer Science Department, Bryn Mawr, PA

[2] University of Maryland, Baltimore County, Department of Computer Science & Electrical Engineering, Baltimore, MD

[3] Lockheed Martin Advanced Technology Laboratories, Artificial Intelligence Research Group, Cherry Hill, NJ

**Abstract.** Multi-view learning algorithms typically assume a complete bipartite mapping between the different views in order to exchange information during the learning process. However, many applications provide only a partial mapping between the views, creating a challenge for current methods. To address this problem, we propose a multi-view algorithm based on constrained clustering that can operate with an incomplete mapping. Given a set of pairwise constraints in each view, our approach propagates these constraints using a local similarity measure to those instances that can be mapped to the other views, allowing the propagated constraints to be transferred across views via the partial mapping. It uses co-EM to iteratively estimate the propagation within each view based on the current clustering model, transfer the constraints across views, and then update the clustering model. By alternating the learning process between views, this approach produces a unified clustering model that is consistent with all views. We show that this approach significantly improves clustering performance over several other methods for transferring constraints and allows multi-view clustering to be reliably applied when given a limited mapping between the views. Our evaluation reveals that the propagated constraints have high precision with respect to the true clusters in the data, explaining their benefit to clustering performance in both single- and multi-view learning scenarios.

## 1. Introduction

Using multiple different views often has a synergistic effect on learning, improving the performance of the resulting model beyond learning from a single view. Multi-view learning is especially relevant to applications that simultaneously collect data from different modalities, with each unique modality providing one or more views of the data. For example, a textual field report may have associated image and video content, and an Internet web page may contain both text and audio. Each view contains unique complementary information about an object; only in combination do the views yield a complete representation of the original object. Concepts that are challenging to learn in one view (e.g., identifying images of patrons at an Italian restaurant) may be easier to recognize in another view (e.g., via the associated textual caption), providing an avenue to improve learning. Multi-view learning can share learning progress in a single view to improve learning in the other views via the direct correspondences between views.

Current multi-view algorithms typically assume that there is a complete bipartite mapping between instances in the different views to represent these correspondences, denoting that each object is represented in all views. The predictions of a model in one view are transferred via this mapping to instances in the other views, providing additional labeled data to improve learning. However, what happens if we have only a partial mapping between the views, where only a limited number of objects have multi-view representations?

This problem arises in many industrial and military applications, where data from different modalities are often collected, processed, and stored independently by specialized analysts. Consequently, the mapping between instances in the different views is incomplete. Even in situations where the connections between views are recorded, sensor availability and scheduling may result in many isolated instances in the different views. Although it is feasible to identify a partial mapping between the views, the lack of a complete bipartite mapping presents a challenge to most current multi-view learning methods. Without a complete mapping, these methods will be unable to transfer any information involving an isolated instance to the other views.

To address this problem, we propose a method for multi-view learning with an incomplete mapping in the context of constrained clustering. Constrained clustering (Basu et al., 2008) is a class of semi-supervised learning methods that cluster data, subject to a set of hard or soft constraints that specify the relative cluster membership of pairs of instances. These constraints serve as background information for the clustering by specifying instance pairs that belong in either the same cluster (a *must-link* constraint) or different clusters (a *cannot-link* constraint). Given a set of constraints in each view, our approach transfers these constraints to affect learning in the other views. With a complete mapping, each constraint has a direct correspondence in the other views, and therefore can be directly transferred between views using current methods. However, with a partial mapping, these constraints may be between instances that do not have equivalences in the other views, presenting a challenge to multi-view learning, especially when the mapping is very limited.

This article proposes the first multi-view constrained clustering algorithm that considers the use of an incomplete mapping between views. Given an incomplete mapping, our approach propagates the given constraints within each view to pairs of instances that have equivalences in the other views. Since these propagated constraints involve only instances with a mapping to the other views, they can be directly transferred to instances in those other views and affect the clustering. The weight of each propagated constraint is given by its similarity to the original constraint, as measured by a local radial basis weighting function that is based on the current estimate of the clustering. This process is depicted in Figure 1. Our approach uses a variant of co-EM (Nigam and Ghani, 2000) to iteratively estimate the propagation within each view, transfer the constraints across views, and update the clustering model. Our experiments show that using co-EM with constraint propagation provides an effective mechanism for multi-view learning under an incomplete mapping between views, yielding significant improvement over several other mechanisms for transferring constraints across views. We also demonstrate that constraint propagation can improve clustering performance even in single-view scenarios, further demonstrating the precision of the inferred constraints and the utility of constraint propagation.

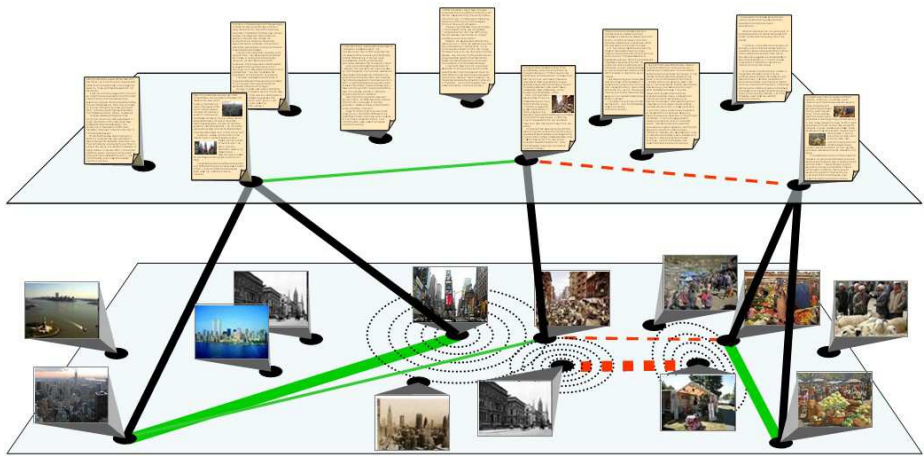

Fig. 1. An illustration of multi-view constrained clustering between two disjoint data views: text and images. We are given a very limited mapping between the views (solid black lines) and a set of pairwise constraints in the images view: two must-link constraints (thick solid green lines) and one cannot-link constraint (thick dashed red line). Based on the current clustering, each given constraint is propagated to pairs of images that are in close proximity to the given constraint and can be mapped to the text view. These propagated must-link and cannot-link constraints (thin solid green and dashed red lines, respectively) are then directly transferred via the mapping to form constraints between text documents and influence the clustering in the next co-EM iteration. (Best viewed in color.)

We first survey related work on constrained clustering and multi-view learning in Section 2, and then present details in Section 3 on the specific constrained clustering and co-EM algorithms on which we base our approach. Section 4 describes our problem setting and mathematical notation. We develop our multi-view constrained clustering algorithm in Section 5, describing the constraint propagation and clustering processes in Sections 5.1–5.2, its extension to more than two views in Section 5.3, and implementation efficiency in Section 5.4. Section 6 evaluates the performance of our approach in several multi-view scenarios (Section 6.3), and then analyzes the performance of constraint propagation inde-pendently through traditional single-view (Section 6.4) clustering. We conclude with a brief discussion of constraint propagation and future work in Section 7.

## 2. Related Work

Our approach combines constrained clustering with multi-view learning. In this section, we briefly review related work on both of these topics.

### 2.1. Constrained clustering

Constrained clustering algorithms (Basu et al., 2008) incorporate side information to influence the resulting clustering model. Most constrained clustering research has focused on using side information given as a set of constraints that specify the relative cluster membership of sets of instances. Typically, these algorithms use both *must-link* constraints, which specify sets of instances that belong in the same cluster, and *cannot-link* constraints, which specify instances that belong in different clusters. Depending on the algorithm, this labeled knowledge may be treated as either hard constraints that cannot be violated, or soft constraints that can be violated with some penalty.

These types of constraints have been successfully integrated into a wide variety of clustering methods (Basu et al., 2008), including K-Means, mixture-models, hierarchical clustering, spectral clustering, and density-based techniques. Although our approach can use most current constrained clustering algorithms, we focus on using K-Means variants and so concentrate our survey on these methods. COP-Kmeans (Wagstaff et al., 2001; Wagstaff, 2002), the first constrained clustering algorithm based on K-Means, performs K-Means clustering while ensuring that all constraints are honored in the cluster assignments. PCK-Means (Basu et al., 2004) performs soft constrained clustering by combining the K-Means objective function with penalties for constraint violations. The MPCK-Means algorithm (Bilenko et al., 2004) builds on PCK-Means to learn the distance metrics for each cluster during the clustering process. We use the PCK-Means and MPCK-Means algorithms as the base clustering methods in our experiments, and describe these two algorithms in more detail in Section 3.1.

Kulis et al. (2009) explore the connections between several different formulations of constrained clustering, showing that semi-supervised weighted kernel K-Means, graph clustering, and spectral clustering are closely related. Based on these connections, they develop a kernel-based constrained clustering approach that can operate on either vector- or graph-based data, unifying these two areas. Domeniconi et al. (2011) propose an alternative formulation of semi-supervised clustering using composite kernels that is suitable for heterogeneous data fusion.

The models generated by PCK-Means/MPCK-Means are equivalent to particular forms of Gaussian mixtures (Bilenko et al., 2004; Basu et al., 2002), and so our work is also closely related to research in constrained mixture modeling. Shental et al. (2004) incorporate hard constraints into Gaussian mixture models using an equivalence set formulation, and Lu et al. (2005) learn a soft constrained Gaussian mixture model using constraints to influence the prior distribution of instances to clusters. More recent work has focused on incorporating constraints into nonparametric mixture models (Vlachos et al., 2009; Mallapragada, 2010).

Due to the intuitive nature of must-link and cannot-link constraints for user interaction, constrained clustering has been applied to the problem of interactive clustering, where the system and user collaborate to generate the model. Cohn et al. (2009) and desJardins et al. (2008) present interactive approaches in which a user iteratively provides feedback to improve the quality of a proposed clustering. In both of these cases, the user feedback is incorporated in the form of constraints. This interactive process is a useful extension that could permit user knowledge to be brought into a multi-view clustering algorithm. Additionally, as we discuss in Section 7, a user could specify these constraints in one view of the data where interaction is quick and intuitive (such as images). Our multi-view clustering algorithm could then be used to automatically propagate and transfer these constraints to affect the clustering of other views where interaction may be more difficult (such as text or other data modalities). Interaction could be further improved by the use of active learning to query the users for specific constraints (Wang and Davidson, 2010; Zhao et al., 2012).

## 2.2. Multi-view learning

Multi-view learning was originated by Blum and Mitchell (1998) in the co-training algorithm for semi-supervised classification. Co-training uses the model for each view to incrementally label the unlabeled data. Labels that are predicted with high confidence are transferred to the corresponding unlabeled instances in the other views to improve learning, and the process iterates until all instances are labeled. Co-training assumes independence between the views, and shows decreased performance when this assumption is violated (Nigam and Ghani, 2000). Dasgupta et al. (2001) provide a PAC generalization analysis of the co-training algorithm that bounds the error of co-training based on the observed disagreement between the partial rules. This analysis relies on the independence of the views, which is often violated in real-world domains.

As an alternative to co-training, Nigam and Ghani (2000) present the co-EM algorithm, an iterative multi-view form of expectation maximization (EM). The co-EM algorithm probabilistically labels all data and transfers those labels to the other view each iteration, repeating this process until convergence. Unlike co-training, it does not require independence between the views in order to perform well. The co-EM algorithm forms the foundation for our approach, and is described in detail in Section 3.2.

Clustering with multiple views was introduced by Bickel and Scheffer (2004), who developed a multi-view EM algorithm that alternates between the views used to learn the model parameters and estimate the cluster assignments. The typical goal of multi-view clustering is to learn models that exhibit agreement across multiple views of the data. Chaudhuri et al. (2009) develop a multi-view approach to support clustering in high-dimensional spaces. They use canonical correlation analysis to construct low-dimensional embeddings from multiple views of the data, then apply co-training to simultaneously cluster the data in these multiple lower-dimensional spaces. de Sa (2005) extends spectral clustering to a multi-view scenario, in which the objective is to minimize the disagreement between views.

In Kumar and Daumé's work, an EM co-training approach is used to perform spectral clustering within one view and use this as a constraint on the similarity graph in another view (Kumar and Daumé, 2011). In subsequent work, they used a regularization approach to optimize the shared clustering (Kumar et al., 2011). In effect, these approaches are propagating the *entire* clustering across views, in contrast to our method, which only propagates the explicit constraints to nearby pairs of instances. Also, unlike our approach, these works assume a complete bipartite mapping between views. Other multi-view clustering variations are based on cross-modal clustering between perceptual channels (Coen, 2005) and information-theoretic frameworks (Sridharan and Kakade, 2008; Gao et al., 2007; Tang et al., 2009).

Recently, several researchers have studied ways to develop mappings between alternative views. Harel and Mannor (2011) describe a method for learning from what they refer to as multiple *outlooks*. Outlooks are similar to views in that they each have different feature spaces; however, there is no assumption that instances would appear in multiple outlooks. (Harel and Mannor mention the possibility of shared instances but do not present any results.) They learn an affine mapping that scales and rotates a given outlook into another outlook by matching the moments of the empirical distributions within the outlooks. This work assumes that the outlooks can be mapped to each other globally with a single affine mapping, whereas our work assumes only local, potentially nonlinear mappings between the views, based on the learned clustering structure. Quadrianto and Lampert (2011) introduce a technique for projecting multiple views into a common, shared feature space, producing a joint distance function across the views. Their work assumes that each instance appears in every view. They use a neighborhood relationship that defines "similar" instances to optimize a similarity function in the shared feature space that respects the neighborhood relationships. The problem setting in their work is rather different from ours, since they assume a complete bipartite mapping between views, and the provided input consists of neighborhood sets rather than pairwise constraints.

There has been a significant amount of recent research on applying co-training to perform clustering in relational data. These methods generally use the attributes of the objects in a relational network as one view, and the relations as another view (or as multiple views). Greene and Cunningham (2009), for example, combine text similarity (the attribute view) with co-citation data (the relational view) to identify communities in citation data. Banerjee et al. (2007) introduce a general framework for co-clustering of attributes and relations, using an approach that finds the clustering with the minimal information loss. Banerjee et al.'s framework can be applied with any single or combination of Bregman loss functions, permitting it to be applied in a variety of different contexts.

Bhattacharya and Getoor (2009) show how the problem of entity resolution (identifying objects in a relational network that refer to the same entity) can be seen as a constrained clustering problem. The way in which information is combined by equating multiple objects in entity resolution can be seen as analogous to multi-view clustering, where each role that the entity plays in the relational network can be thought of as a "view" on that object.

Other applications of multi-view clustering include image search (Chi et al., 2007), biomedical data analysis (Kailing et al., 2004), audio-visual speech and gesture analysis (Christoudias et al., 2008), multilingual document clustering (Kim et al., 2010), word sense disambiguation (Yarowsky, 1995), and e-mail classification (Kiritchenko and Matwin, 2001).

## 3. Background

In this section, we present background on the basic methods on which our approach builds: the PCK-Means and MPKC-Means constrained clustering methods, and the co-EM algorithm.

### 3.1. Constrained clustering methods

Constrained clustering algorithms take as input a set of constraints $\mathcal{C}$ to inform the clustering process. A pairwise constraint $\langle x_i, x_j, w, type \rangle \in \mathcal{C}$ denotes the relative clustering of instances $x_i$ and $x_j$, where the non-negative weight of the constraint is given by $w \in \mathbb{R}_0^+$ (the set of non-negative real numbers) and $type \in \{must\text{-}link, cannot\text{-}link\}$ specifies whether $x_i$ and $x_j$ belong in either the same cluster (a must-link constraint) or different clusters (a cannot-link constraint). In soft constrained clustering, $w$ can be viewed as the penalty for violating the constraint. Throughout this article, wherever the weight or type of constraint are obvious from context, we will omit them and indicate a pairwise constraint as simply $\langle x_i, x_j \rangle$ or $\langle x_i, x_j, w \rangle$. For convenience, we refer to the sets of all must-link and cannot-link constraints as, respectively, $\mathcal{C}_{ml}$ and $\mathcal{C}_{cl}$.

As mentioned earlier, our approach can be combined with any constrained clustering method. Our current implementation supports the PCK-Means (Basu et al., 2004) and MPCK-Means (Bilenko et al., 2004) algorithms; we give results for both methods in Section 6. In the remainder of this section, we provide a brief overview of these methods; further details are available in the original papers.

We first describe the MPCK-Means algorithm, and then show the simplifications that yield PCK-Means. The MPCK-Means algorithm generates a $k$-partitioning of the data $X \subseteq \mathbb{R}^d$ by minimizing the following objective function, which combines the K-Means model with penalties for violating must-link and cannot-link constraints:

$$\mathcal{J}_{MPCK} = \sum_{x_i \in X} \left( \|x_i - \mu_{x_i}\|^2_{\mathbf{M}_{x_i}} - log(det(\mathbf{M}_{x_i})) \right)$$
$$+ \sum_{\langle x_i, x_j, w \rangle \in \mathcal{C}_{ml}} w f_{ml}(x_i, x_j) \mathbb{1}(\mu_{x_i} \neq \mu_{x_j}) \qquad (1)$$
$$+ \sum_{\langle x_i, x_j, w \rangle \in \mathcal{C}_{cl}} w f_{cl}(x_i, x_j) \mathbb{1}(\mu_{x_i} = \mu_{x_j}) \;,$$

where

$$f_{ml}(x_i, x_j) = \tfrac{1}{2}\|x_i - x_j\|^2_{\mathbf{M}_{x_i}} + \tfrac{1}{2}\|x_i - x_j\|^2_{\mathbf{M}_{x_j}} \qquad (2)$$
$$f_{cl}(x_i, x_j) = \|x'_{x_i} - x''_{x_i}\|^2_{\mathbf{M}_{x_i}} - \|x_i - x_j\|^2_{\mathbf{M}_{x_i}} \;, \qquad (3)$$

$\mu_{x_i}$ and $\mathbf{M}_{x_i}$ are respectively the centroid and metric of the cluster to which $x_i$ belongs, $x'_{x_i}$ and $x''_{x_i}$ are the points with the greatest separation according to the $\mathbf{M}_{x_i}$ metric, the function $\mathbb{1}(b) = 1$ if predicate $b$ is true and 0 otherwise, and $\|x_i - x_j\|_{\mathbf{M}} = \sqrt{(x_i - x_j)^\mathsf{T} \mathbf{M} (x_i - x_j)}$ is the Mahalanobis distance between $x_i$ and $x_j$ using the metric $\mathbf{M}$. The first term of $\mathcal{J}_{MPCK}$ attempts to maximize the log-likelihood of the K-Means clustering, while the second and third terms incorporate the costs of violating constraints in $\mathcal{C}$.

---
**Algorithm 1** The co-EM algorithm of Nigam and Ghani (2000)

---
**Input:** first view $X^A$ of the data, second view $X^B$ of the data, and
           an incomplete vector of labels $Y$ over the data.
1: Initialize $\tilde{Y} = Y$.
2: **repeat**
3:     **for** $V \in \{A, B\}$ **do**
4:         Learn model $M^V$ using the data $X^V$ with labels $\tilde{Y}$.       // M-step
5:         Use $M^V$ to label $X^V$, obtaining predicted labels $Y^V$.      // E-step
6:         Set $\tilde{Y}_i = \begin{cases} Y_i & \text{if label } Y_i \text{ is provided} \\ Y_i^V & \text{otherwise} \end{cases}$.
7:     **end for**
8: **until** $M^A$ and $M^B$ have both internally converged

**Output:** The two models $M^A$ and $M^B$.

---

MPCK-Means uses expectation maximization (EM) to locally minimize the objective function $\mathcal{J}_{MPCK}$ to generate the clustering. The E-step consists of assigning each point to the cluster that minimizes $\mathcal{J}_{MPCK}$ from the perspective of that data point, given the previous assignments of points to clusters. The M-step consists of two parts: re-estimating the cluster centroids given the E-step cluster assignments, and updating the metric matrices $\{\mathbf{M}_h\}_{h=1}^{K}$ to decrease $\mathcal{J}_{MPCK}$. The latter step enables MPCK-Means to learn the metrics for each cluster in combination with learning the constrained clustering model. Learning a Mahalanobis metric has also been considered by Xing et al. (2003) and Bar-Hillel et al. (2005). The PCK-Means algorithm is a simplified form of this approach that minimizes the same objective function as MPCK-Means, but eliminates the metric learning aspect and assumes an identity distance metric, setting $f_{ml}(x_i, x_j) = f_{cl}(x_i, x_j) = 1$.

### 3.2. The co-EM algorithm

Co-EM (Nigam and Ghani, 2000) is an iterative algorithm based on expectation maximization that learns a model from multiple views of data. At each iteration, co-EM estimates the model for a view and uses it to probabilistically label all of the data; these labels are then transferred to train another view during the next iteration. Co-EM repeats this process until the models for all views converge. The co-EM algorithm of Nigam and Ghani is given as Algorithm 1; note that this algorithm assumes a complete bipartite mapping between the two views in order to transfer the labels.

Unlike co-training, co-EM does not require the views to be independent in order to perform well. The co-EM algorithm also learns from a large set of data with noisy labels each iteration, in contrast to co-training, which adds few labeled instances to the training set at each iteration. For this reason, Nigam and Ghani (2000) argue that the co-EM algorithm is closer in spirit to the original framework for multi-view learning described by Blum and Mitchell (1998) than the co-training algorithm. The approach we explore in this article uses a vari-ant of co-EM to iteratively infer constraints in each view and to transfer those constraints to affect learning in the other views.

## 4. Preliminaries

Our multi-view constrained clustering approach (described in the next section) takes as input multiple views of the data $\mathcal{X} = \{X^A, X^B, \ldots\}$. Each view $V$ of the data is given by a set of instances $X^V = \{x_1^V, x_2^V, \ldots, x_{n_V}^V\}$, with each $x_i^V \in \mathbb{R}^{d_V}$. The feature set and dimensionality $d_V$ may differ between the views. We will initially focus on the case of two views, given by $X^A$ and $X^B$, and extend our approach to handle an arbitrary number of views in Section 5.3.

Within $\mathcal{X}$, there are pairs of instances that correspond to different views of the same objects. We denote this connection between two instances $x_u^A$ and $x_v^B$ in different views by a relation $r_i = \langle x_u^A, x_v^B \rangle \in X^A \times X^B$. The set of relations $\mathcal{R}^{A \times B} = \{r_1, r_2, \ldots\} \subseteq X^A \times X^B$ defines a bipartite graph between $X^A$ and $X^B$. Most other work on multi-view learning (Nigam and Ghani, 2000; Blum and Mitchell, 1998; Bickel and Scheffer, 2004; Chaudhuri et al., 2009; Kumar and Daumé, 2011) assumes that $\mathcal{R}^{A \times B}$ defines a complete bipartite mapping between the two views. We relax this assumption and consider the case in which $\mathcal{R}^{A \times B}$ provides only a partial mapping between the views. We also consider situations where the partial mapping is extremely limited, with many more data instances than relations between views (i.e., $|\mathcal{R}^{A \times B}| \ll \min(|X^A|, |X^B|)$).

We also have a set of pairwise must-link and cannot-link constraints for each view $V$, given by $\mathcal{C}^V \subset \mathbb{C}^V$, where $\mathbb{C}^V = X^V \times X^V \times \mathbb{R}_0^+ \times \{\textit{must-link}, \textit{cannot-link}\}$ denotes the space of all possible pairwise constraints in view $V$. Depending on the application, these constraints may be either manually specified by the user or extracted automatically from labeled data. Note that the constraints describe relationships between instances within a *single* view, while the mapping $\mathcal{R}^{A \times B}$ defines connections between instances in *different* views.

## 5. Multi-View Constrained Clustering

Our multi-view constrained clustering approach takes as input multiple views of the data $\mathcal{X} = \{X^A, X^B, \ldots\}$, their associated sets of pairwise constraints $\mathcal{C}^A, \mathcal{C}^B, \ldots$, and a (potentially incomplete) mapping $\mathcal{R}^{U \times V}$ between each pair of different views $U$ and $V$. Although we focus primarily on the case of two views $A$ and $B$, we also generalize our approach to multiple views, as described in Section 5.3. The objective of our approach is to determine a $k$-partitioning of the data for each view that respects both the constraints within each view and the mapping between the views.

Our approach, given as Algorithm 2, iteratively clusters each view, infers new constraints within each view, and transfers those inferred constraints across views via the mapping. Through this process, progress in learning the model for one view will be rapidly transmitted to other views, making this approach particularly suited for problems where different aspects of the model are easy to learn in one view but difficult to learn in others.

The base constrained clustering algorithm is given by the *CKmeans* subfunction, which computes the clustering that maximizes the log-likelihood of the data

**Algorithm 2** Multi-view Constrained Clustering with Constraint Propagation

**Input:** first view $X^A$ and constraints $\mathcal{C}^A$,
second view $X^B$ and constraints $\mathcal{C}^B$,
thresholds $t_A \in (0,1]$ and $t_B \in (0,1]$,
the set of cross-view relations $\mathcal{R}^{A \times B}$, and
the number of clusters $k$.

1: Compute the transitive closure of $\mathcal{C}^A \bigcup \mathcal{C}^B \bigcup \mathcal{R}^{A \times B}$.
2: Augment $\mathcal{C}^A$, $\mathcal{C}^B$, and $\mathcal{R}^{A \times B}$ with additional constraints from the transitive closure involving only instances from, respectively, $X^A \times X^A$, $X^B \times X^B$, and $X^A \times X^B$.
3: Let $\hat{X}^A \subseteq X^A$ be the set of instances from $X^A$ involved in $\mathcal{R}^{A \times B}$; similarly define $\hat{X}^B \subseteq X^B$.
4: Define constraint mapping functions $f_{A \mapsto B}$ and $f_{B \mapsto A}$ across views via the set of cross-view relations $\mathcal{R}^{A \times B}$.
5: Initialize the sets of propagated constraints $\mathcal{P}^V = \emptyset$ for $V \in \{A, B\}$.
6: **repeat**
7:  **for** $V \in \{A, B\}$ **do**
8:   Let $U$ denote the opposite view from $V$.

   // M-step
9:   Define the unified set of constraints, mapped with respect to view $V$:
   $$\tilde{\mathcal{C}}^V = \mathcal{C}^V \overset{\max}{\bigcup} f_{U \mapsto V}(\mathcal{P}^U)$$
10:   Update the clustering using constrained K-Means:
   $(P^V, \mathcal{M}^V) = CKmeans(X^V, \tilde{\mathcal{C}}^V, k)$

   // E-step
11:   Estimate the set of propagated constraints:
   $$\mathcal{P}^V = \Big\{ \langle x_i^V, x_j^V \rangle \ : \ x_i^V, x_j^V \in \hat{X}^V \ \wedge$$
   $$\langle x_u^V, x_v^V \rangle \in \mathcal{C}^V \ \wedge$$
   $$W\big(\langle x_i^V, x_j^V \rangle, \langle x_u^V, x_v^V \rangle\big) \geq t_V \Big\}$$
12:  **end for**
13: **until** $P^A$ and $P^B$ have both internally converged

**Output:** the clustering $P^A$ and $P^B$.

---

**Subfunction:** $(P, \mathcal{M}) = CKmeans(X, \mathcal{C}, k)$
*Function prototype for constrained K-Means.*

**Input:** data $X$, must-link and cannot-link constraints $\mathcal{C}$,
and the number of clusters $k$.
**Output:** the partitioning $P$ and set of metrics for each
cluster $\mathcal{M} = \{\mathbf{M}_1, \ldots, \mathbf{M}_k\}$.

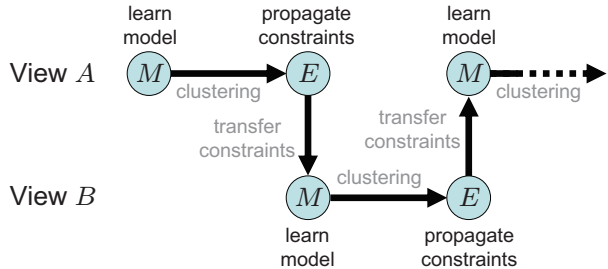

Fig. 2. The relationship between the E-steps and M-steps in the different views.

$X$ given the set of must- and cannot-link constraints $\mathcal{C}$. Our implementation uses either the PCK-Means or MPCK-Means algorithms as the *CKmeans* subfunction due to their native support for soft constraints and, for MPCK-Means, metric learning. However, our approach can utilize other constrained K-Means clustering algorithms, provided that they meet the criteria for the *CKmeans* function listed above.

We fit the clustering model across both views using a variant of the co-EM algorithm (Nigam and Ghani, 2000), described in Section 3.2. In the E-step, we propagate the set of given constraints based on the current clustering models to those instances ($\hat{X}^A$ and $\hat{X}^B$) with direct mappings to the other views (Step 11, further described in Section 5.1). These propagated constraints can then be directly transferred to the other views via the mapping $\mathcal{R}^{A \times B}$ (Steps 4–9, further described in Section 5.2) to influence clustering during the M-step (Step 10). Note that instead of taking the direct union of all of the constraints, we keep only the maximally weighted constraint of each type (must-link and cannot-link) for every pair of instances; this operation is notated by the $\overset{\max}{\cup}$ operator in Step 9.

Following previous work on co-EM and multi-view clustering (Nigam and Ghani, 2000; Bickel and Scheffer, 2004), we iterate the E-step in one view to propagate the constraints followed by the M-step in the other view to transfer those constraints and update the clustering. Each iteration of the co-EM loop (Steps 6–13) contains two iterations of both the E-step and the M-step, one for each view. The relationship between these steps is illustrated in Figure 2. The co-EM process continues until each view has internally converged. We assume that convergence has occurred when the PCK-Means/MPCK-Means objective function's value differs by less than $\epsilon = 10^{-6}$ between successive iterations. Like Nigam and Ghani (2000), we observed that our co-EM variant converged in very few iterations in practice. The iterative exchange of constraints between the views ensures a consistent clustering that respects both the constraints within and the mapping between views. The next two sections describe each step of the co-EM process in detail.

### 5.1. E-step: Constraint propagation

In our model, the sets of pairwise constraints are the sole mechanisms for guiding the resulting clustering. We can directly map a constraint $\langle x_u, x_v \rangle$ between views only if the mapping is defined in $\mathcal{R}^{A \times B}$ for both endpoints $x_u$ and $x_v$ of the constraint. When $\mathcal{R}^{A \times B}$ is incomplete, the number of constraints with such a direct mapping for both endpoints is likely to be small. Consequently, we will be unable to directly map many of the constraints between views; each constraint that we cannot map represents lost information that may have improved the clustering.

Let $\hat{X}^V \subseteq X^V$ be the set of instances for view $V$ that are mapped to another view. Given the initial constraints in $\mathcal{C}^V$, we infer new constraints between pairs of instances in $\hat{X}^V$ based on their local similarity to constraints in $\mathcal{C}^V$. We define this local similarity metric based on the current clustering model for view $V$, and propagate a constraint $\langle x_u^V, x_v^V \rangle \in \mathcal{C}^V$ to a pair of points $x_i^V, x_j^V \in \hat{X}^V$ if the pair is sufficiently similar to the original constraint. This process essentially considers these as spatial constraints (Klein et al., 2002) that affect not only the endpoints, but local neighborhoods of the instance space around those endpoints. Any effect on a pair of points in the neighborhood can be realized as a weighted constraint between those instances. Our constraint propagation method infers these constraints between instances in $\hat{X}^V$ with respect to the current clustering model. Since this set of new constraints (which we refer to as *propagated constraints*) is between instances with a direct mapping to other views, these constraints can be directly transferred to those other views via the mapping $\mathcal{R}^{A \times B}$. This approach can also be interpreted as inferring two weighted must-link constraints $\langle x_u^V, x_i^V \rangle$ and $\langle x_v^V, x_j^V \rangle$ and taking the transitive closure of them with $\langle x_u^V, x_v^V \rangle$ to obtain $\langle x_i^V, x_j^V \rangle$.

The propagation process occurs with respect to the current clustering model for view $V$. Since we use K-Means variants as the base learning algorithm, the learned model is essentially equivalent to a Gaussian mixture model, under particular assumptions of uniform mixture priors and conditional distributions based on the set of constraints (Bilenko et al., 2004; Basu et al., 2002). Therefore, we can consider that each cluster $h$ is generated by a Gaussian with a covariance matrix $\mathbf{\Sigma}_h$. For base clustering algorithms that support metric learning (e.g., MPCK-Means), the cluster covariance is related to the inverse of the cluster metric $\mathbf{M}_h$ learned as part of the clustering process. Bar-Hillel et al. (2005) note that, in practice, metric learning typically constructs the metric modulo a scale factor $\alpha_h$. Although this scale factor does not affect clustering, since only relative distances are required, constraint propagation requires absolute distances. Therefore, we must rescale the learned covariance matrix $\mathbf{M}_h^{-1}$ by $\alpha_h$ to match the data.

We compute $\alpha_h$ based on the empirical covariance $\tilde{\mathbf{\Sigma}}_h$ of the data $P_h \subset X^V$ assigned to cluster $h$, given by

$$\tilde{\mathbf{\Sigma}}_h = \frac{1}{|P_h|} \sum_{x \in P_h} (x - \mu_h)(x - \mu_h)^\mathsf{T} + \gamma \mathbf{I} \ , \tag{4}$$

adding a small amount of regularization $\gamma \mathbf{I}$ to ensure that $\tilde{\mathbf{\Sigma}}_h$ is non-singular for small data samples. Given $\mathbf{M}_h^{-1}$ and $\tilde{\mathbf{\Sigma}}_h$, we compute $\alpha_h$ as the scale such that the variances of the first principal component of each matrix are identical. We take the eigendecomposition of each matrix

$$\mathbf{M}_h^{-1} = \mathbf{Q}_{\mathbf{M}_h^{-1}} \mathbf{\Lambda}_{\mathbf{M}_h^{-1}} \mathbf{Q}_{\mathbf{M}_h^{-1}}^\mathsf{T} \qquad\qquad \tilde{\mathbf{\Sigma}}_h = \mathbf{Q}_{\tilde{\mathbf{\Sigma}}_h} \mathbf{\Lambda}_{\tilde{\mathbf{\Sigma}}_h} \mathbf{Q}_{\tilde{\mathbf{\Sigma}}_h}^\mathsf{T} \tag{5}$$

to yield diagonal matrices of eigenvalues in $\mathbf{\Lambda}_{\mathbf{M}_h^{-1}}$ and $\mathbf{\Lambda}_{\tilde{\mathbf{\Sigma}}_h}$. To derive the scale factor $\alpha_h$, we ensure that both first principal components have equal variances,

which occurs when

$$\alpha_h = \frac{\max\left(\mathbf{\Lambda}_{\tilde{\mathbf{\Sigma}}_h}\right)}{\max\left(\mathbf{\Lambda}_{\mathbf{M}_h^{-1}}\right)} \ , \tag{6}$$

yielding $\mathbf{\Sigma}_h = \alpha_h \mathbf{M}_h^{-1}$ as the covariance matrix for cluster $h$. When the base learning algorithm does not support metric learning, as with PCK-Means, we can instead use $\mathbf{\Sigma}_h = \tilde{\mathbf{\Sigma}}_h$ as cluster $h$'s covariance matrix. The model for cluster $h$ is then given by

$$G_h(x^V) = \exp\left(-\tfrac{1}{2}\left\|x^V - \mu_h^V\right\|_{\mathbf{\Sigma}_h^{-1}}^2\right) \ , \tag{7}$$

where

$$\left\|x^V - \mu_h^V\right\|_{\mathbf{\Sigma}_h^{-1}}^2 = (x^V - \mu_h^V)^\mathsf{T} \mathbf{\Sigma}_h^{-1} (x^V - \mu_h^V) \tag{8}$$

is the squared Mahalanobis distance between $x^V$ and $\mu_h^V$ according to the cluster's rescaled metric $\mathbf{\Sigma}_h^{-1}$.

We assume that each constraint should be propagated with respect to the current clustering model, with the shape (i.e., covariance) of the propagation being equivalent to the shape of the respective clusters (as given by their covariance matrices). Additionally, we assume that the propagation distance should be proportional to the constraint's location in the cluster. Intuitively, a constraint located near the center of a cluster can be propagated a far distance, up to the cluster's edges, since being located near the center of the cluster implies that the model has high confidence in the relationship depicted by the constraint. Similarly, a constraint located near the edges of a cluster should only be propagated a short distance, since the relative cluster membership of these points is less certain at the cluster's fringe.

We propagate a given constraint $\langle x_u^V, x_v^V \rangle \in \mathcal{C}^V$ to two other points $x_i^V, x_j^V \in X^V$ according to a Gaussian radial basis function (RBF) of the distance as $\langle x_i^V, x_j^V \rangle$ moves away from $\langle x_u^V, x_v^V \rangle$. Under this construction, the weight of the propagated constraint decreases according to the RBF centered in $2d_V$-dimensional space at the original constraint's endpoints $\begin{bmatrix} x_u^V & x_v^V \end{bmatrix} \in \mathbb{R}^{2d_V}$ with a covariance matrix $\mathbf{\Sigma}_{uv}^V$ based on the respective clusters' covariance matrices.

To form the propagation covariance matrices for each endpoint, we scale the covariance matrix associated with endpoint $x_u^V$ by the weight assigned to that endpoint according to the clustering model (Equation 7). This ensures that the amount of propagation falls off with increasing distance from the centroid, in direct relation to the model's confidence in the cluster membership of $x_u^V$. The covariance matrix for the constraint propagation function is then given by

$$\mathbf{\Sigma}_{uv}^V = \begin{bmatrix} G_{c_u}(x_u^V)\,\mathbf{\Sigma}_{c_u} & \mathbf{0} \\ \mathbf{0} & G_{c_v}(x_v^V)\,\mathbf{\Sigma}_{c_v} \end{bmatrix} \ , \tag{9}$$

where $c_u$ denotes the cluster of $x_u$ and $\mathbf{0}$ denotes the $d_V \times d_V$ zero matrix. This construction assumes independence between $x_u^V$ and $x_v^V$. While this assumption is likely to be violated in practice, we empirically show that it yields good results. For convenience, we represent the covariance matrices associated with each endpoint by $\mathbf{\Sigma}_{x_u^V} = G_{c_u}(x_u^V)\,\mathbf{\Sigma}_{c_u}^V$ for $x_u^V$ and $\mathbf{\Sigma}_{x_v^V} = G_{c_v}(x_v^V)\,\mathbf{\Sigma}_{c_v}^V$ for $x_v^V$. Figure 3 illustrates the results of this process on an example cluster.

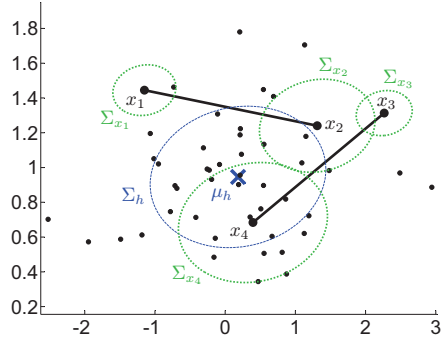

Fig. 3. Constraint propagation applied to a single example cluster $h$, showing the learned covariance matrix $\mathbf{\Sigma}_h$ (dashed blue ellipse) rescaled to fit the data, two constraints (solid black lines), and the weighting functions centered at each endpoint (dotted green ellipses), which decrease in variance as they move farther from the centroid $\mu_h$.

Given a constraint $\langle x_u^V, x_v^V, w, \textit{type} \rangle \in \mathcal{C}^V$ and two candidate points $x_i^V \in X^V$ and $x_j^V \in X^V$, we can now estimate the weight of the propagated constraint $\langle x_i^V, x_j^V \rangle$ as

$$W\big(\langle x_i^V, x_j^V \rangle, \langle x_u^V, x_v^V \rangle\big) = w \times \max\Big( W'\big(\langle x_i^V, x_j^V \rangle, \langle x_u^V, x_v^V \rangle\big), \\ W'\big(\langle x_j^V, x_i^V \rangle, \langle x_u^V, x_v^V \rangle\big) \Big) \quad (10)$$

where

$$W'\big(\langle x_i^V, x_j^V \rangle, \langle x_u^V, x_v^V \rangle\big) = \exp\left(-\tfrac{1}{2} \left\| x_i^V - x_u^V \right\|^2_{\mathbf{\Sigma}_{x_u^V}^{-1}}\right) \times \\ \exp\left(-\tfrac{1}{2} \left\| x_j^V - x_v^V \right\|^2_{\mathbf{\Sigma}_{x_v^V}^{-1}}\right) . \quad (11)$$

Since the ordering of the instances matters in the propagation, we compute both possible pairings of constraint endpoints ($x_u^V$ and $x_v^V$) to target endpoints ($x_i^V$ and $x_j^V$), taking the maximum value of the propagation in Equation 10 to determine the best match. Under this propagation scheme, a constraint propagated to its own endpoints is given a weight of $w$ (since the second term of the RHS of Equation 10 will be 1); the weight of the propagated constraint decreases as the endpoints $x_i^V$ and $x_j^V$ move farther from $x_u^V$ and $x_v^V$. Section 5.4 describes mechanisms for implementing constraint propagation efficiently, taking advantage of the independence assumption between the two endpoints of a constraint and memoization of repeated computations.

The E-step of Algorithm 2 (Step 11) uses Equation 10 to propagate all given constraints within each view to those instances $\hat{X}^V$ with cross-view mappings, thereby inferring the expected value of constraints between those instances given the current clustering. Using this set of expected constraints $\mathcal{P}^V$, we then update the current clustering model in the M-step, as described in the next section.

### 5.2. M-step: Updating the clustering model

Given the expected constraints $\mathcal{P}^U$ between instances in $\hat{X}^U$, we transfer those constraints to the other views and then update the clustering model to reflect these new constraints. These steps together constitute the M-step of Algorithm 2.

Any propagated constraint where both endpoints are in $\hat{X}^U$ can be transferred to another view $V$ via the bipartite mapping $\mathcal{R}^{U \times V}$. We define a mapping function $f_{U \mapsto V} : \mathbb{C}^U \mapsto \mathbb{C}^V$ that takes a given constraint $\langle x_i^U, x_j^U, w, type \rangle \in \mathcal{P}^U$ and maps it to constrain instances in $X^V$ by:

$$f_{U \mapsto V}(\langle x_i^U, x_j^U, w, type \rangle) = \{\langle x_u^V, x_v^V, w, type \rangle : \langle x_i^U, x_u^V \rangle \in \mathcal{R}^{U \times V} \wedge \\ \langle x_j^U, x_v^V \rangle \in \mathcal{R}^{U \times V}\} \;. \quad (12)$$

Using this construction, we can define the mapping functions $f_{B \mapsto A}$ and $f_{A \mapsto B}$ in Algorithm 2. We then use these functions $f_{A \mapsto B}$ and $f_{B \mapsto A}$ to map propagated constraints between views in Step 9, transferring constraints inferred in one view to the other related views. These transferred constraints (from view $U$) can then be combined with the original constraints in each view $V$ to inform the clustering. Instead of taking the direct union of these constraints, we keep only the maximally weighted constraint between each pair of instances to form the set

$$\tilde{\mathcal{C}}^V = \mathcal{C}^V \bigcup^{\max} f_{U \mapsto V}(\mathcal{P}^U) \;, \quad (13)$$

since each inferred constraint represents an estimate of the minimal strength of the pairwise relationship.

The partitioning for view $V$ can then be computed by clustering the data $X^V$ subject to the constraints in $\tilde{\mathcal{C}}^V$ (Step 10). The *CKmeans* subfunction computes the clustering that maximizes the log-likelihood of the data subject to the set of constraints, thereby completing the M-step of Algorithm 2.

### 5.3. Extension to multiple views

Algorithm 2 can be easily extended to support more than two views. Each view $X^V$ independently maintains its own sets of given constraints $\mathcal{C}^V$, threshold $t_V$, data $\hat{X}^V \subseteq X^V$ involved in any cross-view relations, current partitioning $P^V$, current cluster metrics $\mathcal{M}^V$, and propagated constraints $\mathcal{P}^V$. To handle more than two views, we maintain separate mappings $\mathcal{R}^{U \times V}$ for each pair of views $X^U$ and $X^V$ and use each mapping to define pairwise mapping functions $f_{U \mapsto V}$ and $f_{V \mapsto U}$ between views. For $D$ views, $X^{(1)}, \ldots, X^{(D)}$, this approach will yield $D^2 - D$ mapping functions.

To generalize our approach to more than two views, we hold each set of propagated constraints fixed, and iteratively update the clustering (M-step), then recompute the set of propagated constraints (E-step) for one view. The unified sets of constraints for each view $V$ becomes (Step 9):

$$\tilde{\mathcal{C}}^V = \mathcal{C}^V \bigcup_{U=1}^{\max D} f_{U \mapsto V}(\mathcal{P}^U) \;, \quad (14)$$

under the convention that $f_{U \mapsto U}(\mathcal{P}^U) = \emptyset$. Each iteration of co-EM loops over the E-steps and M-steps for all views, and proceeds until the clustering for each view converges. The order of the views may either be fixed or chosen randomly at each iteration.

### 5.4. Implementation efficiency

The overall computational complexity of Algorithm 2 is determined by the maximum number of EM iterations and by the complexity of the *CKMeans* function, which depends on the chosen clustering algorithm. Besides these aspects, the constraint propagation step (Step 11) incurs the greatest computational cost. To make this step computationally efficient, our implementation relies on the independence assumption inherent in Equation 11 between the two endpoints of the constraint. To efficiently compute the weight of all propagated constraints, we memoize the value of each endpoint's propagation

$$G(x_i^V, x_u^V) = \exp\left(-\frac{1}{2}(x_i^V - x_u^V)^\mathsf{T} \mathbf{\Sigma}_{x_u^V}^{-1}(x_i^V - x_u^V)\right) \qquad (15)$$

for $x_i^V \in \hat{X}^V$ and $x_u^V \in \bar{X}^V$, where $\bar{X}^V$ is the set of points involved in $\mathcal{C}^V$. Through memoization, we reduce the constraint propagation step to $|\hat{X}^V| \times |\bar{X}^V|$ Gaussian evaluations. Memoization applies similarly to all other views. Each constraint propagation is inherently independent from the others, making this approach suitable for parallel implementation using Hadoop/MapReduce (Dean and Ghemawat, 2008).

When the covariance matrix $\mathbf{\Sigma}_{x_u^V}$ is diagonal, we can further reduce the computational cost through early stopping of the Gaussian evaluation once we are certain that the endpoint's propagation weight will be below the given threshold $t_V$. When $\mathbf{\Sigma}_{x_u^V}$ is diagonal, given by $\mathbf{\Sigma}_{x_u^V} = diag(\sigma_1^2, \sigma_2^2, \ldots, \sigma_{d_V}^2)$,

$$G(x_i^V, x_u^V) = \exp\left(-\frac{1}{2}\sum_{k=1}^{d_V} \frac{(x_{i,k}^V - x_{u,k}^V)^2}{\sigma_k^2}\right) \quad . \qquad (16)$$

Since a constraint is only propagated when the weight exceeds $t_V > 0$ and the maximum propagation for each Gaussian weight $G(x_i^V, x_u^V) \in [0,1]$, we only need to evaluate $W'(\langle x_i^V, x_j^V\rangle, \langle x_u^V, x_v^V\rangle)$ when both $G(x_i^V, x_u^V) \geq t_V$ and $G(x_j^V, x_v^V) \geq t_V$. Therefore, we must ensure that

$$t_V \leq \exp\left(-\frac{1}{2}\sum_{k=1}^{d_V} \frac{(x_{i,k}^V - x_{u,k}^V)^2}{\sigma_k^2}\right) \qquad (17)$$

$$-2\ln t_V \geq \sum_{k=1}^{d_V} \frac{(x_{i,k}^V - x_{u,k}^V)^2}{\sigma_k^2}. \qquad (18)$$

Since all terms in the RHS summation are positive, we can compute them incrementally and stop early once the sum exceeds $-2\ln t_V$, since we will never need to evaluate any propagation weight $W'(\cdot)$ involving $G(x_i^V, x_u^V)$. In our implementation, we set $G(x_i^V, x_u^V) = 0$ in any cases where we can guarantee that $G(x_i^V, x_u^V) < t_V$.

## 6. Evaluation

We evaluated multi-view constrained clustering on a variety of data sets, both synthetic and real, showing that our approach improves multi-view learning under an incomplete mapping as compared to several other methods. Our results reveal that the constraints inferred by propagation have high precision with respect to the true clusters in the data. We also examined the performance of Constraint Propagation in the individual views, revealing that Constraint Propagation can also improve performance in single-view clustering scenarios.

### 6.1. Data sets

In order to examine the performance of our approach under various data distributions, we use a combination of synthetic and real data in our experiments. We follow the methodology of Nigam and Ghani (2000) to create these multi-view data sets by pairing classes together to create "super-instances" consisting of one instance from each class in the pair. The two original instances then represent two different views of the super-instance, and their connection forms a mapping between the views. This methodology can be trivially extended to an arbitrary number of views. These data sets are described below and summarized in Table 1.

**Four Quadrants** is a synthetic data set composed of 200 instances drawn from four Gaussians in $\mathbb{R}^2$ space with identity covariance. The Gaussians are centered at the coordinates $(\pm 3, \pm 3)$, one in each of the four quadrants. Quadrants I and IV belong to the same cluster and quadrants II and III belong to the same cluster. The challenge in this simple data set is to identify these clusters automatically, which requires the use of constraints to improve performance beyond random chance. To form the two views, we drew 50 instances from each of the four Gaussians, divided them evenly between views, and created mappings between nearest neighbors that were in the same quadrant but different views.

**Protein** includes 116 instances divided among six classes of proteins, denoted $\{c_1, c_2, \ldots, c_6\}$. This data set was previously used by Xing et al. (2003). To create multiple views of this data set, we partition it into two views containing respectively instances from classes $\{c_1, c_2, c_3\}$ and $\{c_4, c_5, c_6\}$. We connect instances between the following pairs of classes to create the two views: $c_1$ & $c_4$, $c_2$ & $c_5$, and $c_3$ & $c_6$. Through this construction, a model learned for clustering $\{c_1, c_2, c_3\}$ in one view can be used to inform the clustering of $\{c_4, c_5, c_6\}$ in the other view. Since the clusters do not contain the same numbers of instances, some instances within each view are isolated in the mapping.

**Letters/Digits** uses the letters-IJL and digits-389 data sets previously used by Bilenko et al. (2004). These are subsets of the letters and digits data sets from the UCI machine learning repository (Asuncion and Newman, 2007) containing only the letters $\{I, J, L\}$ and the digits $\{3, 8, 9\}$, respectively. We map instances between views according to the following pairings: $I$ & 3, $J$ & 8, and $L$ & 9, leaving those instances without a correspondence in the other view isolated in the mapping.

| Name | $n_A$ | $n_B$ | #dims | k | $t_V$ |
|---|---|---|---|---|---|
| Four Quadrants | 200 | 200 | 2 | 2 | 0.75 |
| Protein | 67 | 49 | 20 | 3 | 0.50 |
| Letters/Digits | 227 | 317 | 16 | 3 | 0.95 |
| Rec/Talk | 100 | 94 | 50 | 2 | 0.75 |

Table 1. Properties of each data set and the values of all parameters used in the experiments.

**Rec/Talk** is a subset of the 20 Newsgroups data set (Rennie, 2003), containing 5% of the instances from the newsgroups {*rec.autos*, *rec.motorcycles*} in the *rec* view, and 5% of the instances from the newsgroups {*talk.politics.guns*, *talk.politics.mideast*} in the *talk* view. We process each view independently, removing stop words and representing the data as a binary vector of the 50 most discriminatory words as determined by Weka's string-to-wordvector filter (Witten and Frank, 2005). As in the previous data sets, we form the mapping between views by pairing clusters in order.

We create a low-dimensional embedding of each data set using the spectral features (Ng et al., 2001) in order to improve clustering, with the exception of Four Quadrants, for which we use the original features because the dimensionality is already low. For each view $V$, we compute the pairwise affinity matrix $\mathbf{A}$ between the instances $x_i$ and $x_j$ using a radial basis function of their distance, given by $A_{i,j} = \exp(-||x_i - x_j||^2/2\sigma^2)$. We use $\sigma = 1$ as the rate at which the affinity falls off with increasing distance. From $\mathbf{A}$, we form the normalized Laplacian matrix (Chung, 1994) for the data set, given by $\mathcal{L} = \mathbf{I} - \mathbf{D}^{-\frac{1}{2}} \mathbf{A} \mathbf{D}^{-\frac{1}{2}}$, where $\mathbf{D}$ is the diagonal degree matrix $D_{i,i} = \sum_{j=1}^{n_V} A_{i,j}$ and $\mathbf{I}$ is the identity matrix. The eigendecomposition of the normalized Laplacian matrix $\mathcal{L} = \mathbf{Q}\mathbf{\Lambda}\mathbf{Q}^\mathsf{T}$ yields the spectral features for the data set in the columns of the eigenvector matrix $\mathbf{Q}$. We keep the $2^{nd}$ through $d+1^{th}$ eigenvectors (corresponding to the $2^{nd}$ through $d+1^{th}$ lowest eigenvalues in $\mathbf{\Lambda}$) as the features for clustering; we discard the first eigenvector since it is constant and therefore does not discriminate between the instances. In this article, we use $d = \lceil\sqrt{d_V}\rceil$ for Protein and Letters/Digits, and $d = 5$ for the Rec/Talk data set. Additionally, we standardize all features to have zero mean and unit variance. These spectral features are computed independently between the different views, further emphasizing that the mapping is the only connection between views.

### 6.2. Methodology

Within each view, we use the true cluster labels on the instances to sample a set of pairwise constraints, ensuring equal proportions of must-link and cannot-link constraints. The weight of all sampled constraints $w$ is set to 1. We also sample a portion of the mapping to use for transferring constraints between views. Both the sets of constraints and the mapping between views are resampled each trial.

We compare **Constraint Propagation** against several other potential methods for transferring constraints:

**Direct Mapping** transfers only those constraints that already exist between instances in $\hat{X}^V$. This approach is equivalent to other methods for multi-view learning that are only capable of transferring labeled information if there is a direct mapping between views.

**Cluster Membership** can be used to infer constraints between instances in $\hat{X}^V$. This approach simply considers the relative cluster membership for each pair of instances in $\hat{X}^V$ and infers the appropriate type of constraint with a weight of 1. It represents the direct application of co-EM to infer and transfer constraints between views via the partial mapping.

**Single View** performs constrained clustering on each of the individual views in isolation and serves as a lower baseline for the experiments.

For the base constrained clustering algorithms, we use the PCK-Means and MPCK-Means implementations provided in the WekaUT extension[1] to the Weka machine learning toolkit (Witten and Frank, 2005) with their default values. For PCK-Means, Constraint Propagation uses the full empirical covariance for each cluster; for MPCK-Means, it uses the diagonal weighted Euclidean metrics that are learned on a per-cluster basis by MPCK-Means.

We measure performance using the pairwise F-measure – a version of the information-theoretic F-measure adapted to measure the number of same-cluster pairs for clustering (Basu, 2005). The pairwise F-measure is the harmonic mean of precision and recall, given by

$$F\text{-}measure = \frac{2 \cdot precision \cdot recall}{precision + recall} \tag{19}$$

$$precision = \frac{|\mathcal{P}_{correct}|}{|\mathcal{P}_{pred}|} \qquad recall = \frac{|\mathcal{P}_{correct}|}{|\mathcal{P}_{same}|} \ ,$$

where $\mathcal{P}_{pred}$ is the set of entity pairs predicted to be in the same community, $\mathcal{P}_{same}$ is the set of entity pairs actually in the same community, and $\mathcal{P}_{correct} = \mathcal{P}_{pred} \bigcap \mathcal{P}_{same}$ is the set of correct predictions. We take the mean of the F-measure performance for all views, yielding a single performance measure for each experiment.

In each trial, we consider performance as we vary the number of constraints used for learning and the percentage of instances in each view that are mapped to the other views. Our results are shown in Figure 4, averaged over 100 trials.

### 6.3. Discussion of multi-view clustering results

As shown in Figure 4, Constraint Propagation clearly performs better than the baseline of Single View clustering, and better than Cluster Membership for inferring constraints in all cases, except for when learning with few constraints on Letters/Digits. Constraint Propagation also yields an improvement over Direct Mapping for each percentage of instances mapped between views, as shown in Figure 5. Unlike Direct Mapping, Constraint Propagation is able to transfer those constraints that would otherwise be discarded, increasing the performance of multi-view clustering. The performance of both Constraint Propagation and Direct Mapping improve as the mapping becomes more complete between the views, with Constraint Propagation still retaining an advantage over Direct Mapping even with a complete mapping, as shown in all data sets. We hypothesize that in the case of a complete mapping, Constraint Propagation behaves similarly to spatial constraints (Klein et al., 2002), warping the underlying space with the inference of new constraints that improve performance.

---

[1] http://www.cs.utexas.edu/users/ml/risc/code/

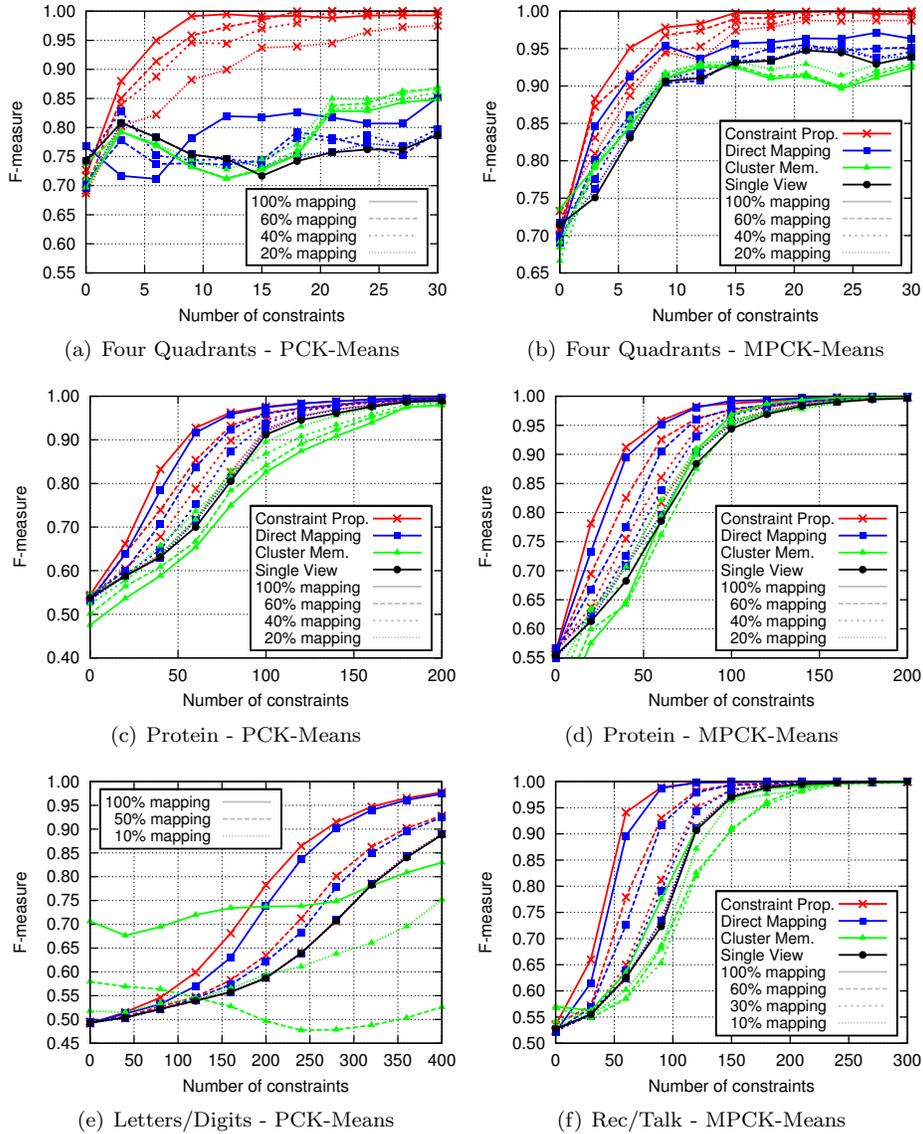

**Fig. 4.** Comparison of multi-view constrained clustering performance. The percentage of instances mapped between views (e.g., 20%, 40%, 100%) is denoted by the type of the line (dotted, dashed, solid), and the constraint transfer method is denoted by the color and marker shape. In several plots, we truncated the key due to space limitations; those plots use the same markers and colors to depict the constraint transfer methods as the other plots. (Best viewed in color.)

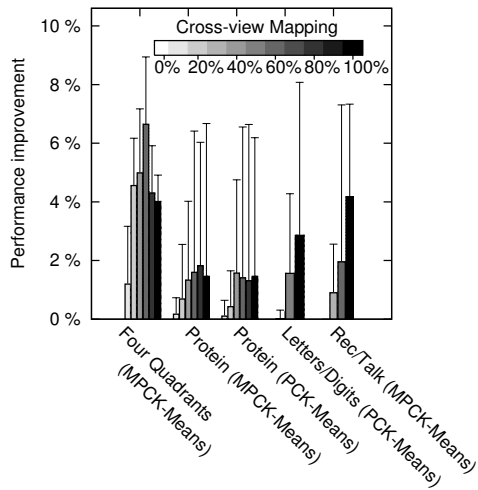

Fig. 5. The performance improvement of Constraint Propagation over Direct Mapping in Figure 4, averaged over the learning curve. The peak whiskers depict the maximum percentage improvement.[2]

On these data sets, the number of constraints inferred by Constraint Propagation is approximately linear in the number of original constraints, as shown in Figure 6. Clearly, as the mapping between views becomes more complete, Constraint Propagation is able to infer a larger number of constraints between those instances in $\hat{X}^V$.

The improvement in clustering performance is due to the high precision of the propagated constraints. Figure 7 shows the average weighted precision of the propagated constraints for the 100% mapping case, measured against the complete set of pairwise constraints that can be inferred from the true cluster labels. The proportion that each propagated constraint contributed to the weighted precision is given by the constraint's inferred weight $w$. We also measured the precision of propagated constraints for various partial mappings, and the results were comparable to those for the complete mapping. The constraints inferred through propagation show a high average precision of 98–100% for all data sets, signifying that the propagation method infers very few incorrect constraints.

Interestingly, the constraint propagation method works slightly better for cannot-link constraints than for must-link constraints. This phenomenon can be explained by a counting argument that there are many more chances for a cannot-link constraint to be correctly propagated than a must-link constraint. For example, with $k$ clusters where each cluster contains $n/k$ instances, each given must-link constraint can be correctly propagated to $numMLprop = \binom{n/k}{2} - 1$ other pairs of instances in the same cluster. However, each given cannot-link constraint can be correctly propagated to $numCLprop = \binom{n}{2} - k\binom{n/k}{2} - 1$ other pairs of instances that belong in different clusters. The number of propagated cannot-link constraints is computed by taking the number of possible different constraints, $\binom{n}{2} - 1$, and subtracting off the total number of possible must-link constraints, $k\binom{n/k}{2}$. Note that $numCLprop$ is *much* greater than $numMLprop$ (e.g., for $n = 1,000$ and $k = 10$, $numCLprop = 449,999 \gg 4,949 = numMLprop$), implying that a must-link constraint has much less chance of being propagated correctly than a cannot-link constraint.

---

[2] We omit the Four Quadrants (PCK-Means) results from Figure 5 due to the relatively high performance gain of Constraint Propagation, which averages a 21.3% improvement over Direct Mapping with peak gains above 30%. The much greater scale of these improvements are obvious and would make the smaller (but still significant) gains in the other domains less discernible in the figure.

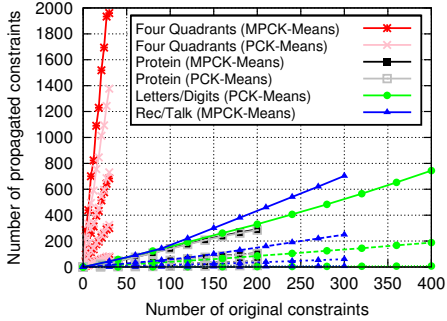 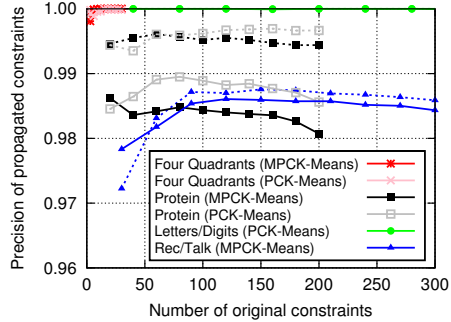

Fig. 6. The number of propagated constraints as a function of the number of original constraints in multi-view clustering, averaged over all co-EM iterations. The line type (solid, dashed, dotted) depicts the mapping percentage, as defined in the corresponding plot in Figure 4. (Best viewed in color.)

Fig. 7. The precision of the propagated must-link (solid line) and cannot-link (dashed line) constraints in multi-view clustering, as measured against the true cluster labels. (Best viewed in color.)

We found that for high-dimensional data, the curse of dimensionality causes instances to be so far separated that Constraint Propagation is only able to infer constraints with a very low weight. Consequently, it works best with a low-dimensional embedding of the data, motivating our use of spectral feature reduction. Other approaches could also be used for creating the low-dimensional embedding, such as principal components analysis or manifold learning.

Additionally, like other multi-view algorithms, we found Constraint Propagation to be somewhat sensitive to the cutoff thresholds $t_V$, but this problem can be remedied by using cross-validation to choose $t_V$. Too high a threshold yields performance identical to Direct Mapping (since no constraints would be inferred), while too low a threshold yields the same decreased performance as exhibited by other co-training algorithms. For this reason, we recommend setting $t_V$ to optimize cross-validated performance over the set of constrained instances.

We ran several additional experiments on multi-view data sets with poor mappings and distributions that violated the mixture-of-Gaussians assumption of K-Means clustering. On these data sets, Constraint Propagation decreased performance in some cases, due to inferring constraints that were not justified by the data. This would occur, for example, in clusters with a nested half-moon shape or concentric rings, where Constraint Propagation would incorrectly infer constraints between instances in the opposing cluster. In these cases, clustering using only the directly mapped constraints yielded the best performance.

### 6.4. Constraint propagation within single views

To dissect the benefits of Constraint Propagation independently from multi-view learning, we evaluated the use of Constraint Propagation in traditional single-view clustering scenarios. We found that using the constraints inferred by propagation can also improve performance in single-view clustering, further explaining its performance in multi-view scenarios and also demonstrating the high quality of the inferred constraints.

Instead of propagating constraints across views, we used Constraint Propagation to infer additional constraints within a single view $A$, employed those constraints to inform the clustering for that view, and then used the resulting clusters to guide the propagation for the next iteration of clustering. This process was repeated until the clustering model for view $A$ converged. To implement this approach, we modified Algorithm 2 to use only a single view by eliminating all steps involving view $B$, altering step 3 to set $\hat{X}^A = X^A$, and altering step 9 to compute the unified set of constraints as $\tilde{\mathcal{C}}^V = \mathcal{C}^V \stackrel{\max}{\cup} \mathcal{P}^V$. With these modifications, the algorithm alternates the M-step clustering and E-step Constraint Propagation to learn a model for the single view $A$. We then evaluated the performance of this single-view clustering method on individual data sets.

Figure 8 depicts the performance of single-view clustering with Constraint Propagation on a sample of the data sets, averaged over 100 trials. All parameters were specified as described in Table 1, with the exceptions of using $t_V = 0.7$ for protein and $t_V = 0.98$ for digits, since we now set $t_V$ for each individual view. We compare single-view clustering with Constraint Propagation against standard constrained clustering. Our analysis omits the other two methods tested in the multi-view experiments, since Direct Mapping has no analogous case for single-view scenarios, and constraints inferred by Cluster Membership in a single view will not alter the resulting model.

These results show that the constraints inferred by propagation within a single view can also be used to improve clustering performance in that view. Both MPCK-Means and PCK-Means show improved performance using the constraints inferred by propagation to augment the original constraints. The maximum improvement from Constraint Propagation occurs with a moderate number of constraints; as we would expect, Constraint Propagation provides little benefit when the number of original constraints is either very small or very large. Even in the single-view case, the number of constraints inferred by propagation is roughly linear in the number of original constraints (Figure 9). We also found that the inferred constraints have very high precision with the true cluster assignments in all single-view scenarios (Figure 10). As in the multi-view experiments, the inferred cannot-link constraints have slightly higher precision than the inferred must-link constraints.

As mentioned in the previous section, we found that Constraint Propagation can decrease performance when applied to data sets that violate the assumptions of K-Means clustering. The two conditions assumed by K-Means that are most necessary for successful Constraint Propagation are that: (1) the clusters are globular under the cluster metrics, and (2) the clusters are separable (i.e., have little overlap). However, it may be difficult to determine *a priori* whether a particular data set meets these conditions, and is therefore an appropriate candidate for Constraint Propagation. For this reason, it is important to determine whether the data set satisfies these conditions during the clustering process; we next discuss two potential methods for making this determination.

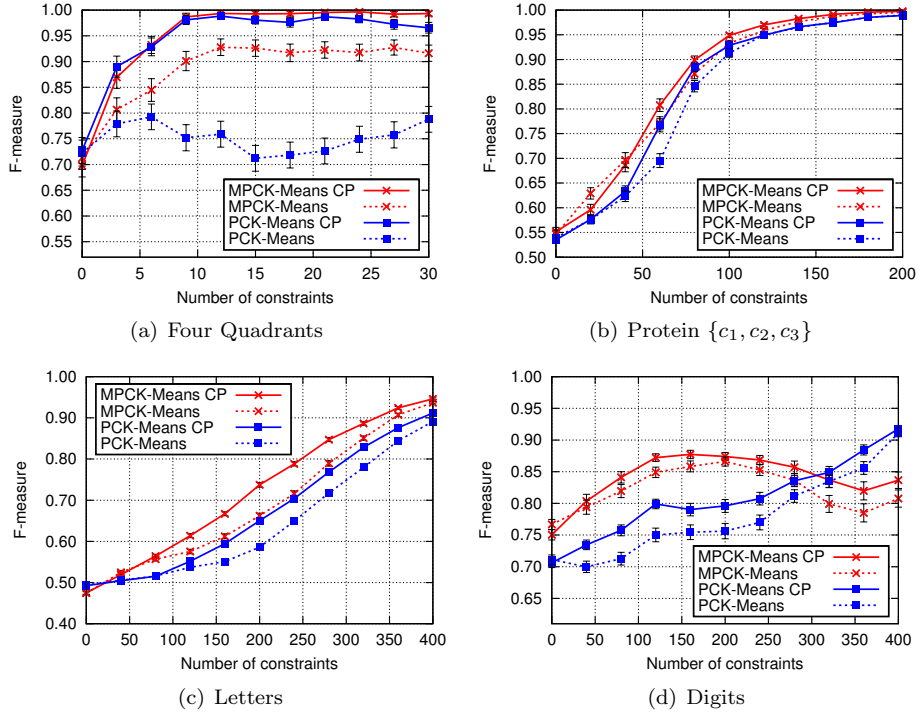

Fig. 8. Comparison of single-view clustering performance with and without Constraint Propagation. The solid lines depict clustering with Constraint Propagation (CP), and the dotted lines depict standard constrained clustering. The base algorithm, either MPCK-Means or PCK-Means, is denoted by the color and marker shape. The black error bars depict the standard error of the mean performance. (Best viewed in color.)

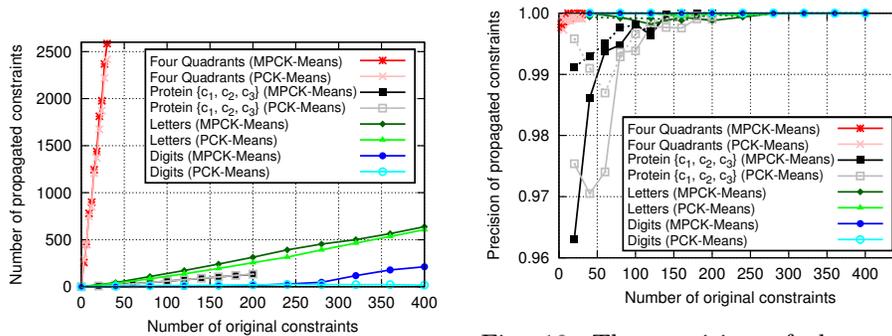

Fig. 9. The number of propagated constraints in single-view clustering as a function of the number of original constraints, averaged over all iterations. (Best viewed in color.)

Fig. 10. The precision of the propagated must-link (solid line) and cannot-link (dashed line) constraints in single-view clustering, as measured against the true cluster labels. (Best viewed in color.)

In cases where the data set does violate these conditions, we have observed that the clustering is primarily driven by the provided constraints rather than the K-Means component. One of the strengths of the MPCK-Means and PCK-Means algorithms is that in situations where K-Means does poorly on the data, large numbers of constraints can overcome the K-Means component and ensure the proper clustering. In terms of the PCK-Means/MPCK-Means objective function (Equation 1), the second and third terms of the equation (those involving the must- and cannot-link constraints) dominate the first term (K-Means) in these situations. When the data set does not meet these conditions, the cluster distributions specified by the learned centroids and metrics have little in common with the resulting partitions. During the clustering process, we can examine the learned cluster distributions for disagreement with the resulting partitions; the presence of a large disagreement is one indication that the data set may be inappropriate for Constraint Propagation.

To determine whether the data set meets the second condition, we can also directly examine the resulting clusters for overlap. Since constraints are propagated based on RBF distance from the given constraint and the relative cluster membership of the endpoints, Constraint Propagation may not infer constraints correctly between points belonging to overlapping clusters. We can detect overlap in the clusters by measuring the KL divergence (Kullback and Leibler, 1951) between their learned distributions; the presence of significant overlap is another indication that the data set may violate the conditions necessary for successful Constraint Propagation. Currently, we simply alert the user if the data set may be inappropriate for Constraint Propagation based on either indicator; we leave the robust determination of whether Constraint Propagation is guaranteed to improve performance to future work.

## 7. Conclusion

Constraint Propagation has the ability to improve multi-view constrained clustering when the mapping between views is incomplete. Besides improving performance, Constraint Propagation also enables information supplied in one view to be propagated and transferred to improve learning in other views. This is especially beneficial for applications in which it is more natural for users to interact with particular views of the data. For example, users may be able to rapidly and intuitively supply constraints between images, but may require a lengthy examination of other views (e.g., text or audio) in order to infer constraints. In other cases, users may not have access to particular views (or even data within a view) due to privacy restrictions. In these scenarios, our approach would be able to propagate user-supplied constraints both within and across views to maximize clustering performance.

Beyond our approach, there are a variety of other methods that could be adapted for learning with a partial mapping between views, such as manifold alignment and transfer learning. Further work on this problem will improve the ability to use isolated instances that do not have a corresponding multi-view representation to improve learning, and enable multi-view learning to be used for a wider variety of applications.

**Acknowledgements.** We would like to thank Kiri Wagstaff, Katherine Guo, Tim Oates, Tim Finin, and the anonymous reviewers for their feedback. This work is based on the first author's Master's thesis at UMBC, and was partially conducted while the first author was at Lockheed Martin Advanced Technology Laboratories. This research was supported by a graduate fellowship from the Goddard Earth Sciences and Technology Center at UMBC, NSF ITR grant #0325329, ONR grant #N00014-10-C-0192, and Lockheed Martin.

## Author Biographies

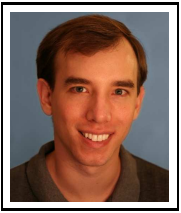

**Eric Eaton** is a Visiting Assistant Professor in the Computer Science Department at Bryn Mawr College. Prior to joining Bryn Mawr, he was a Senior Research Scientist in the Artificial Intelligence Research Group at Lockheed Martin Advanced Technology Laboratories, and a part-time Visiting Assistant Professor at Swarthmore College. Dr. Eaton's research interests are in machine learning and artificial intelligence, including lifelong learning, knowledge transfer, interactive AI, and collaborative knowledge discovery.

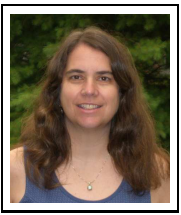

**Marie desJardins** is a Professor in the Department of Computer Science and Electrical Engineering at the University of Maryland, Baltimore County. Prior to joining the faculty in 2001, Dr. desJardins was a Senior Computer Scientist at SRI International in Menlo Park, California. Her research is in artificial intelligence, focusing on the areas of machine learning, multi-agent systems, planning, interactive AI techniques, information management, reasoning with uncertainty, and decision theory.

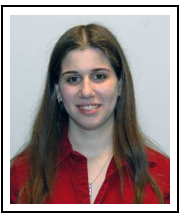

**Sara Jacob** is a Research Software Engineer in the Artificial Intelligence Research Group at Lockheed Martin Advanced Technology Laboratories. She received her B.S. in computer science from Rutgers University. During her time at Rutgers, she helped spearhead the Rutgers Emerging Scholars in Computer Science (RES-CS) program, a peer mentoring program which seeks to encourage more women and minorities to enter the field of computer science.

*Correspondence and offprint requests to*: Eric Eaton, Computer Science Department, Bryn Mawr College, Bryn Mawr, PA 19010, USA. E-mail: eeaton@cs.brynmawr.edu